\documentclass[12pt]{article}
\usepackage[square,numbers]{natbib}
\usepackage[utf8]{inputenc} 
\usepackage[T1]{fontenc} 
\usepackage{hyperref} 
\usepackage{url} 
\usepackage{booktabs} 
\usepackage{amsfonts} 
\usepackage{nicefrac} 
\usepackage{microtype} 
\usepackage{lipsum}
\usepackage{fancyhdr} 
\usepackage{graphicx} 
\graphicspath{{media/}} 
\usepackage{xcolor}
 \usepackage{setspace}
\usepackage[para]{threeparttable}

\usepackage{amsmath}
\usepackage{nccmath}
\usepackage{tabularx}
\usepackage{multirow}
\usepackage{amssymb}
\usepackage{subfig}
\usepackage{enumitem}
\usepackage{multicol}
\usepackage{array}
\usepackage{physics}
\usepackage{authblk}

\usepackage{tcolorbox}
\usepackage{longtable}
\graphicspath{ {images/} }
\providecommand{\keywords}[1]
{
  \small	
  \textbf{\textit{Keywords---}} #1
}

\title{Applying BERT and ChatGPT for Sentiment Analysis of Lyme Disease in Scientific Literature

}
\author[1]{Teo Susnjak}
\affil[1]{Senior Lecturer in Computer Science and IT, 

School of Mathematical and Computational Sciences, 

Massey University, Auckland, New Zealand }

\date{\empty}
\begin{document}
\maketitle

\begin{abstract} 

This chapter presents a practical guide for conducting Sentiment Analysis using Natural Language Processing (NLP) techniques in the domain of tick-borne disease text. The aim is to demonstrate the process of how the presence of bias in the discourse surrounding chronic manifestations of the disease can be evaluated. The goal is to use a dataset of 5643 abstracts collected from scientific journals on the topic of chronic Lyme disease to demonstrate using Python, the steps for conducting sentiment analysis using pre-trained language models and the process of validating the preliminary results using both interpretable machine learning tools, as well as a novel methodology of using emerging state-of-the-art large language models like ChatGPT. This serves as a useful resource for researchers and practitioners interested in using NLP techniques for sentiment analysis in the medical domain.

\end{abstract}


\keywords{Sentiment analysis, Lyme disease text analysis; NLP; BERT, ChatGPT, SHAP, language models, explainable AI }

\maketitle

\section{Introduction}\label{sec1}

Natural Language Processing (NLP) is an effective tool that can be used to extract insights from text. The underlying techniques of NLP can be employed to gather and analyze the content of text, such as topic discovery, and identify patterns and trends. The aim of this chapter is to demonstrate how NLP can be effectively utilised in conjunction with sentiment analysis in the domain of scientific tick-borne disease text. The goal of this chapter is to demonstrate the first steps of the process of determining the presence of bias in the discourse relating to chronic manifestations of the disease, while the general methodology is broadly applicable to other contexts. 

The current consensus amongst academics regarding the existence of chronic Lyme disease is that it is a controversial and debated topic \cite{levesque2019multiple,rebman2020post}. Some scientists and medical professionals believe that chronic Lyme disease, also known as post-treatment Lyme disease syndrome (PTLDS), is a real and debilitating condition that can occur after a person has been infected with the bacteria that cause Lyme disease \cite{wong2022review}. They argue that the symptoms of PTLDS, such as fatigue, pain, and cognitive impairment, can persist for months or even years after initial treatment, and that these symptoms are caused by ongoing infection with the Lyme bacteria.

On the other hand, other scientists and medical professionals argue that there is little scientific evidence to support the existence of chronic Lyme disease \cite{halperin2015chronic}. They claim that the symptoms of PTLDS are not caused by ongoing infection with the Lyme bacteria, but rather by other factors such as immune dysfunction, chronic fatigue syndrome, or depression. They argue that the available evidence does not support the use of prolonged or repeated antibiotic treatment for PTLDS, and that such treatment can be harmful.

This chapter presents a practical guide for conducting sentiment analysis using NLP techniques, with the aim of demonstrating the process of identifying potential bias within medical journals with respect to their position on the chronic nature of tick-borne diseases. To achieve this goal, a dataset comprising 5643 abstracts collected from scientific journals between 2000 and 2021 on the topic of chronic Lyme disease is used. The objective of this chapter is to demonstrate using small amounts of Python code, how preliminary sentiment analysis can be performed using pre-trained language models and validated using explainable AI tools such as SHAP and the emerging state-of-the-art large language models like ChatGPT. The use of these language models in combination with comprehensive datasets allows for a meaningful analysis of the discourse surrounding chronic Lyme disease in the medical community. 
This chapter serves as a resource for researchers and practitioners interested in using NLP techniques for sentiment analysis in the medical domain.

\section{Materials}\label{sec1}

When selecting a tool for NLP, it is useful to take into account one's technical aptitude and willingness to invest time into learning a new tool for this task. The plethora of available tools, each with their own unique features and capabilities, necessitates a thorough examination of the tool's alignment with the individual's specific needs and long-term objectives. 

Code-free tools that solely require point-and-click interaction are available. They can sometimes be restrictive; however, they can be a more suitable pathway for those who are new to NLP and programming as they usually enable a swifter initiation without the need for acquiring a significant background in programming. 

Conversely, for experienced developers who are comfortable with coding, advanced tools such as Python or R with a vast array of libraries and packages for NLP may be more suitable. However, this chapter will demonstrate that one can achieve a significant amount of analysis with minimal Python code, and the option to perform hybrid analyses with multiple technologies that includes both coding and point-and-click software is also a useful approach.

\subsection{Data Collection Tools}

This chapter assumes that the task of data collection has already been performed by the researcher, therefore this process will not be covered. However, several tools are mentioned below for reference in case the reader needs to perform this task or is interested in alternative technologies. 

Scrapy is an open-source web crawling framework that is well-documented and handles a variety of functionalities that can simplify the building of web crawlers using Python. ParseHub is a code-free web scraping tool with high usability and can export the extracted data in a variety of formats. Octoparse is a  cloud-based point-and-click web data extraction solution that can be used by those without programming skills.

\subsection{NLP Tools}

This chapter uses Python for demonstration purposes. Anaconda is a recommended distribution of the Python programming language that includes a collection of libraries and tools for data science which can be installed on personal machines. Getting started can however be much faster with a cloud-based provider which hosts this technology instead. Note \ref{point:installation} briefly describes the installation process and alternative cloud-based solutions. R is equally a suitable programming tool for NLP. A selection of alternative tools for NLP are listed below.

\begin{itemize}

\item KNIME is an open-source data analytics platform that can be used for NLP tasks. It offers a user-friendly, graphical interface that allows users to create, execute, and share workflows for NLP tasks such as text preprocessing, feature extraction, and sentiment analysis. Some advantages of KNIME include its flexibility, ability to handle large datasets, and range of visualisation options with a drag-and-drop interface. However, there is a learning curve to overcome initially and it may not offer as many pre-built models as other NLP tools.

\item OpenNLP is an open-source Java library for NLP that includes tools for tokenisation, POS tagging, named entity recognition, and more. OpenNLP is well-documented, and it has a wide range of NLP tools. OpenNLP may not have the best performance for some tasks, and it may be less user-friendly than other libraries.

\item MonkeyLearn is a cloud-based NLP platform that includes tools for general text analysis, including sentiment analysis. It does not require programming and provides access to pre-built models that can be customised.

\item SAS is a commercial software suite that provides extensive tools for data analysis and capabilities for NLP. It offers features such as text mining, sentiment analysis, and entity recognition. Its ability to integrate with other SAS tools and handle large datasets makes it well-suited for NLP tasks. However, cost and a steep learning curve have to be taken into account.

\end{itemize}

\section{Methods}\label{sec1}

The methodology describes the steps necessary to perform sentiment analysis on abstracts extracted from peer-reviewed journals. The goal is to demonstrate how the first stages of insight extraction from the attitudes and opinions expressed in these abstracts can be conducted in order to better understand how the topic of interest is being discussed in the academic community. Conclusions will not be drawn here, but instead,  researchers will be left with a foundation from which subsequent and more sophisticated analysis can be conducted.

The approach described is specifically tailored to the context of analysing abstracts from peer-reviewed literature, taking into account the unique characteristics and constraints of this type of text. The primary constraint is that academic prose is generally more objective and less susceptible to revealing subjective sentiments. The methods outlined in this section provide an overview of the procedures and techniques used to collect and analyze the data, together with examples of outputs. The overall aim is to provide some guidance, allowing for replication of the preliminary results.

\subsection{Process}
\begin{figure}[hbt]
    \centering
    \includegraphics[scale=0.42]{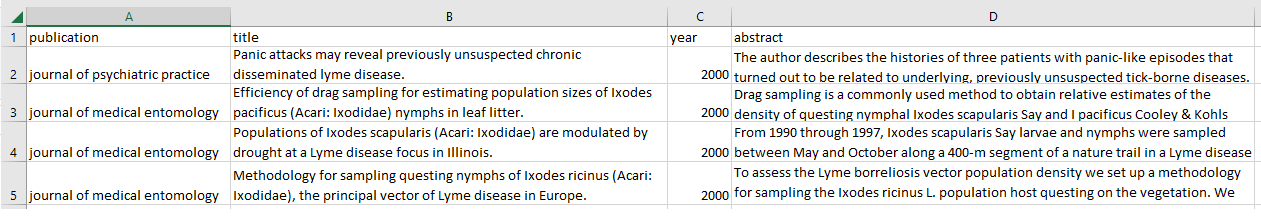}
    \caption{Example of the dataset showing the first five rows and the key columns, being the journal name, paper title, year of publication and the text of the abstract.}
    \label{Fig:dataset}
    \end{figure}

    \paragraph{1. Data Collection:} The first step in performing NLP for sentiment analysis in the domain of Lyme disease is to gather a large dataset of text data. This can include data from sources such as patient and/or physician discussion fora, patient testimonials, social media platforms, and medical journals or other scientific research publications. During the collection process, it is important to filter the collection of data to ensure that the data is specific to Lyme disease and any specific sub-topic as required. Fig. \ref{Fig:dataset} shows an example of the dataset used here, depicting a minimum set of necessary columns needed for the analysis and the structure of the dataset stored in a CSV format.

\begin{figure}[hbt]
    \centering
    \includegraphics[scale=0.33]{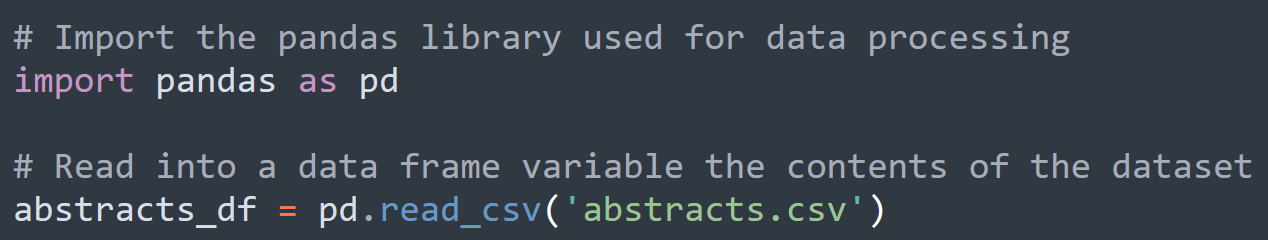}
    \caption{Example of reading the cleaned dataset into a Python variable.}
    \label{Fig:read_dataset}
    \end{figure}

    \paragraph{2. Data Cleaning:} After collecting the data, it is important to clean the dataset to remove any irrelevant information. It can be helpful to correct typographical and grammatical errors if these can be done efficiently or automated\footnote{GPT-series and BERT-like transformer-based models can automate the tasks of correcting typographical and grammatical errors in text.}; however, it is not absolutely necessary for sentiment analysis using the models in this demonstration. Duplicate data should be removed when detected. When scraping data from the web, irrelevant data such as advertisements can be inadvertently collected, and these should be deleted if and when detected. The benefits of data cleaning generally result in improvements in accuracy and robustness of findings; however, the time needed to clean the data comprehensively ought to be weighed with marginal returns that the efforts sometimes return in large datasets. Some level of noise in the datasets can be expected. Once the dataset has been cleaned using a tool of choice, it can be read into a variable using Python for conducting sentiment analysis as seen in Fig. \ref{Fig:read_dataset}.

    \paragraph{3. Sentiment Analysis:} After cleaning the data, sentiment analysis can be performed according to the following steps.

\begin{enumerate}[label=\Roman*.]
\item \label{point:model_selection} Model selection: Many models exist for performing sentiment analysis. Researchers have the option of using widely available machine-learning models or lexicon-based methods. Here, a pre-trained deep-learning transformer model is demonstrated called BERT \cite{devlin2018bert}. BERT (Bidirectional Encoder Representations from Transformers) is a language model developed by Google and one that has been widely used for various NLP tasks \cite{deepa2021bidirectional} (Note \ref{point:domain_mismatch}). The model is customisable, and the particular model used here (details can be found in Note \ref{point:model_details}) has been fine-tuned for sentiment analysis tasks on the business domain. It also does not require some additional NLP processing tasks to be performed before it is used (Note \ref{point:tokenisation}). Fig. \ref{Fig:BERT} demonstrates the code which makes the specific model available for use in a Python script.

    \begin{figure}[hbt]
    \centering
    \includegraphics[scale=0.3]{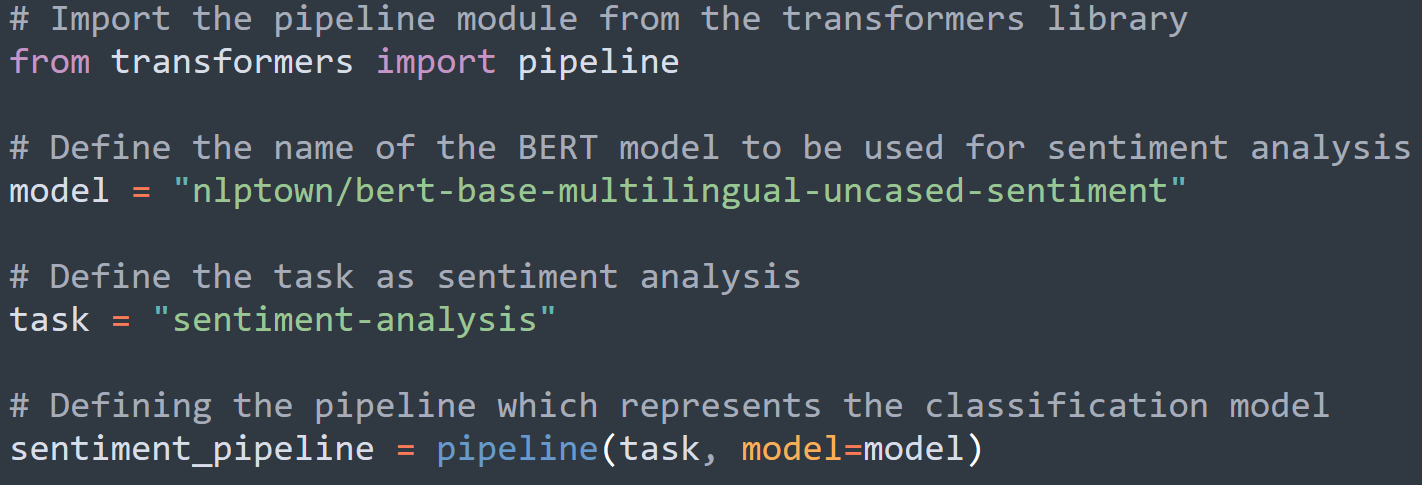}
    \caption{Definition of the specific BERT model with capabilities of performing sentiment analysis.}
    \label{Fig:BERT}
    \end{figure}

\item \label{point:text_classification} Text classification: Once a pre-trained model has been selected and imported into a Python script, it then needs to be applied to each individual text data point (abstract) and the output from the models signifying the score and the sentiment label need to be stored. This particular BERT model outputs a sentiment value which ranges from 0 to 1, as well as a label that represents a sentiment classification ranging from 1 to 5.  Fig. \ref{Fig:apply_bert} demonstrates how the BERT model is applied to classify each abstract and how the sentiment score and label are stored into new columns of the data frame variable.

    \begin{figure}[hbt]
    \centering
    \includegraphics[scale=0.3]{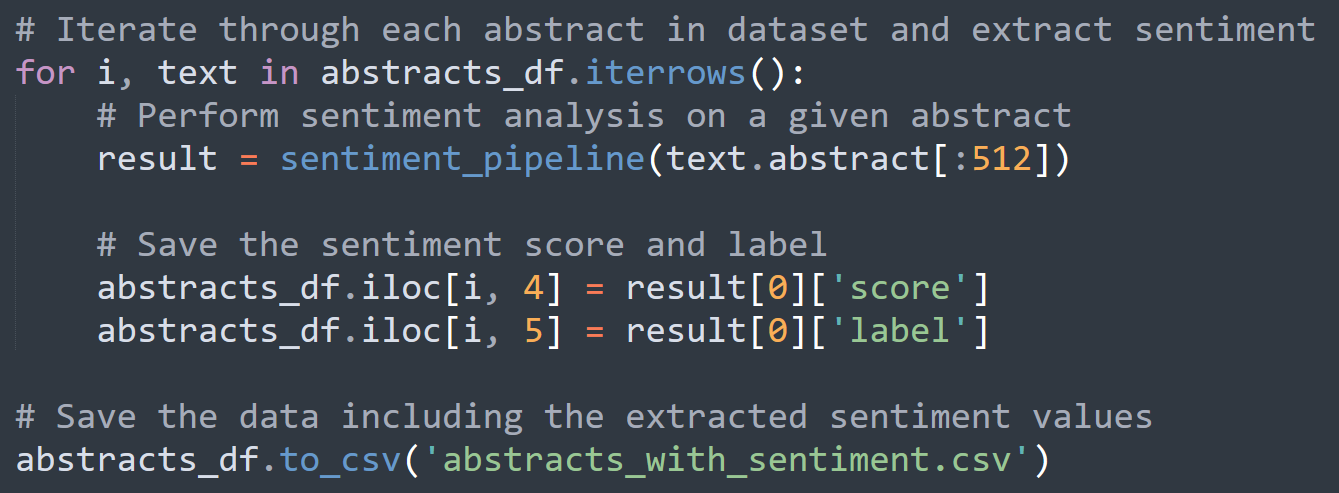}
    \caption{Application of the BERT model to classify sentiment values of each abstract and the extraction of its outputs and saves them in a separate file for subsequent processing and analysis tasks.}
    \label{Fig:apply_bert}
    \end{figure}

\item \label{point:visualisation} Visualisation: After assigning sentiment scores to the entire text corpus, it is important to visually analyse and assess the validity of these scores. The validity and significance of the sentiment scores will depend on the appropriateness of the sentiment analysis model and the specific domain it is being applied to. The following figures provide examples of useful visualisations for this process. Fig. \ref{Fig:hist} illustrates the distribution of sentiment scores across all abstracts in the demonstration dataset.

    \begin{figure}[hbt]
    \includegraphics[scale=0.2]{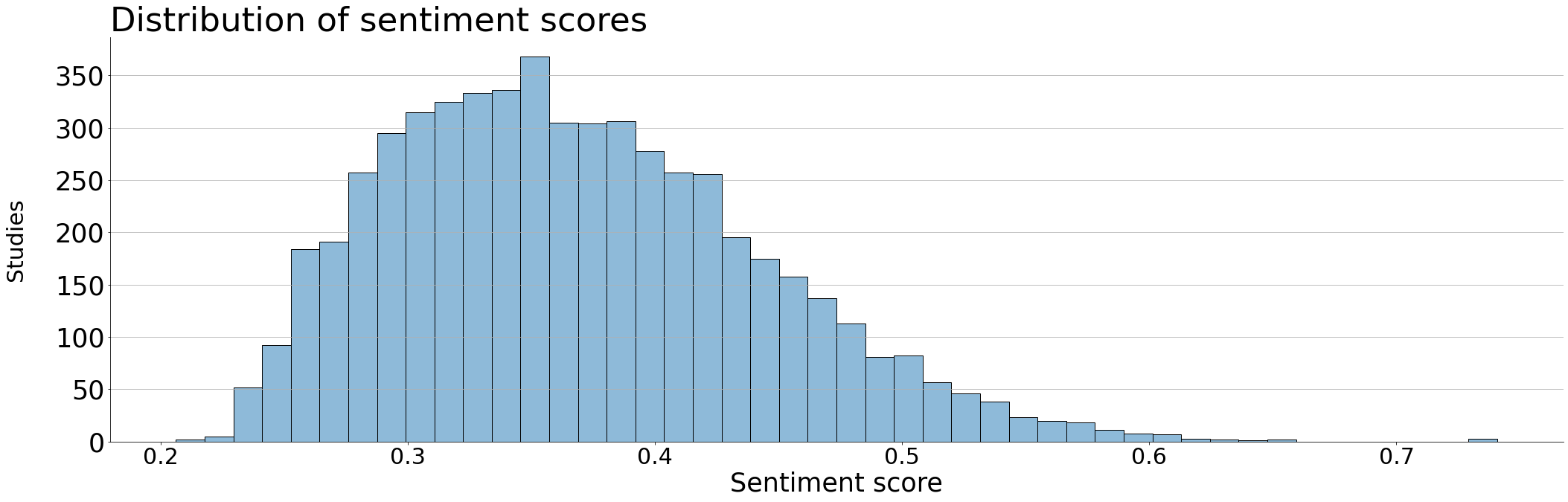}
    \caption{The distribution of sentiment scores across all the abstracts in the dataset.}
    \label{Fig:hist}
    \end{figure}

An additional analysis can involve reviewing the changes in the published sentiment scores of abstracts across time. This is shown in Fig. \ref{Fig:year}.
    
    \begin{figure}[hbt]
    \includegraphics[scale=0.2]{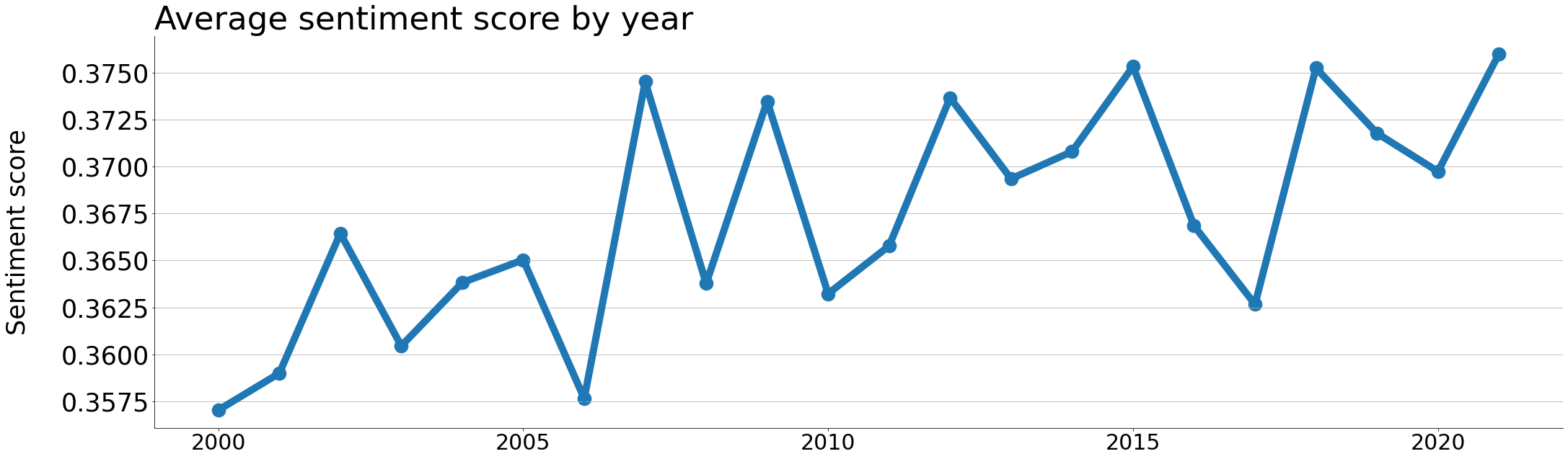}
    \caption{The trend of the average sentiment scores from 2010 to 2020.}
    \label{Fig:year}
    \end{figure}

Sentiments can also be grouped by journals and aggregated in order to highlight if significant differences exist. This can be seen in Fig. \ref{Fig:journal}.

    \begin{figure}[hbt]
    \includegraphics[scale=0.2]{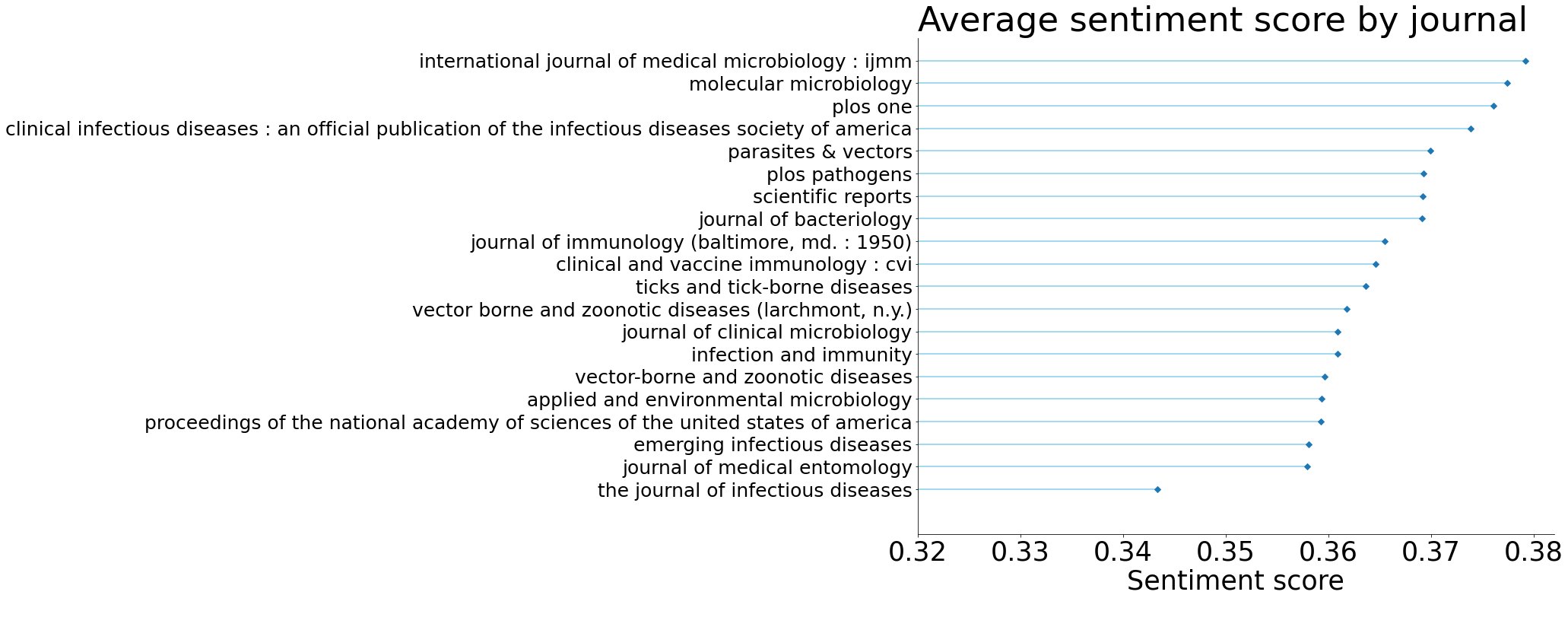}
    \caption{The distribution of average sentiment scores across the top 20 journals by volume of published abstracts from the dataset.}
    \label{Fig:journal}
    \end{figure}
    
\item \label{point:inspecting_classifications} Inspection and Validation of Text: An important step for researchers before conducting more advanced analyses and drawing conclusions from the results of sentiment analysis, is to view a sample selection of text being classified and to examine how the model is interpreting various words in terms of their sentiment scores. This is crucial since models trained on one specific domain are likely to perform sub-optimally on text from another domain given divergences in terminologies and jargon. In this example, a model trained to detect sentiment in customer reviews is being applied to the medical context, where domain-specific words are likely to be associated with some unexpected sentiment values. Fig. \ref{Fig:text} illustrates an example of inspecting the classification of text by word using the SHAP \cite{NIPS2017_7062} tool. The figure shows an example of text that has been classified as having a lower than average overall sentiment score where sentences in red like 'making it difficult to diagnose' strongly confirm its low score, while sentences like 'Prompt treatment can prevent disability.', influencing the overall score towards a higher sentiment designation.

    \begin{figure}[hbt]
    \centering
    \includegraphics[scale=0.52]{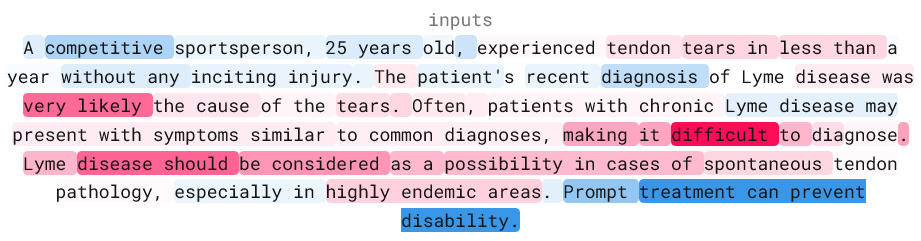}
    \caption{An example of a journal paper abstract for a paper covering chronic Lyme disease. The sample text received an overall sentiment score of 0.31 and a sentiment classification of 2 on a scale from 1 to 5. The figure shows the influence that various words have on the final prediction of the sentiment, with words highlighted in red, positively influencing the prediction towards this classification, while the blue has an opposite effect with darker colours indicating a stronger influence.  }
    \label{Fig:text}
    \end{figure}

The code used in Python to generate Fig. \ref{Fig:text} can be seen in Fig. \ref{Fig:shap_code}. 

    \begin{figure}[hbt]
    \centering
    \includegraphics[scale=0.3]{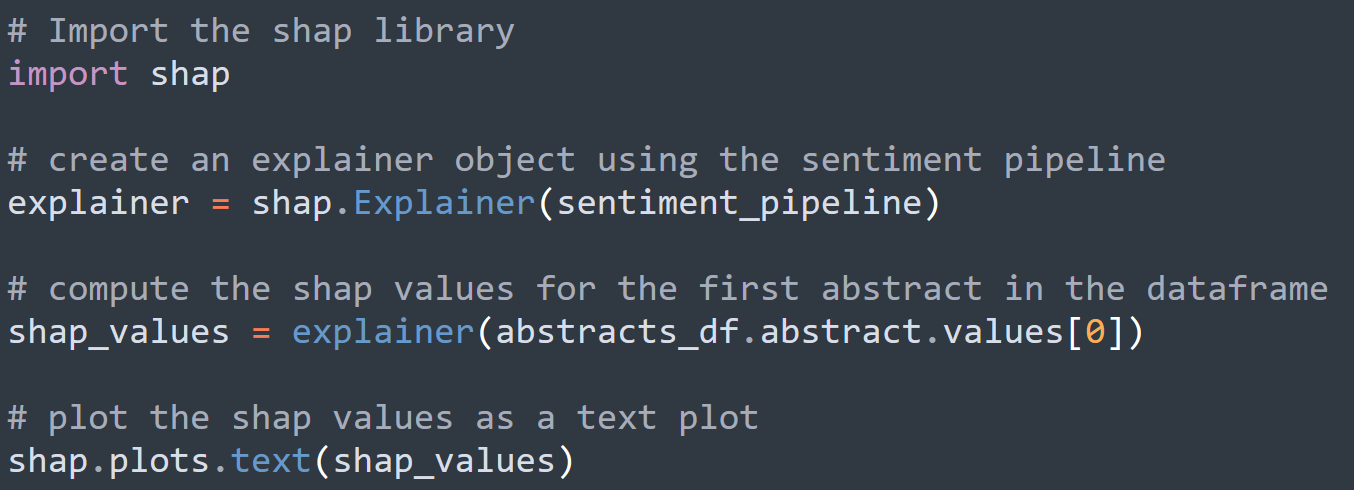}
    \caption{This Python code uses the SHAP library to create an explainer object that can be used to compute SHAP values for the input text (abstracts). The SHAP values are then plotted as a text plot, which shows how each word in the input text contributes to the overall sentiment prediction made by the model. }
    \label{Fig:shap_code}
    \end{figure}

A recently developed AI technology can be used to inspect and validate different samples of text and to cross-check the sentiment and subjectivity interpretation of the text via state-of-the-art Large Language Models (LLMs) like ChatGPT (or GPT-3 via APIs). The following figures provide an example of ChatGPT's interpretation of the given text, including its level of subjectivity (Fig. \ref{Fig:chatgpt_subjectivity}) and sentiment (Fig. \ref{Fig:chatgpt_sentiment}) which agree with the BERT model and the prior analyses.

    \begin{figure}[hbt]
    \centering
    \includegraphics[scale=0.4]{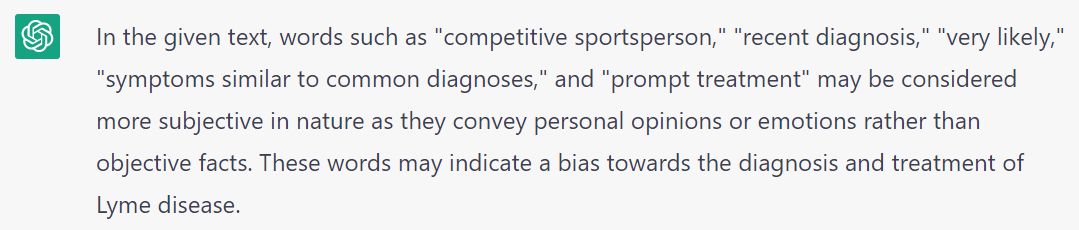}
    \caption{ChatGPT output when asked identify specific words and phrases which have a higher level of subjectivity. }
    \label{Fig:chatgpt_subjectivity}
    \end{figure}

     \begin{figure}[hbt]
    \centering
    \includegraphics[scale=0.4]{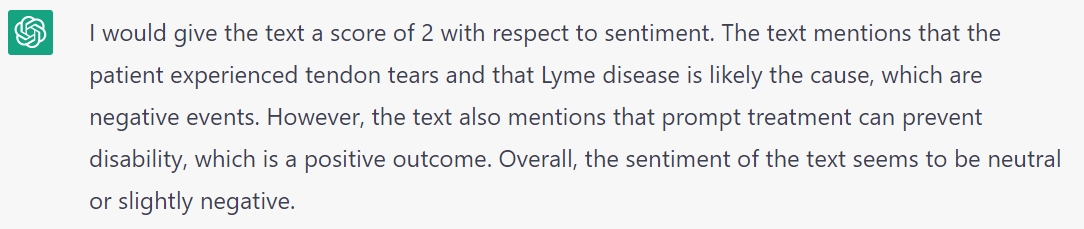}
    \caption{ChatGPT output when asked to classify the abstract for sentiment ranging from 1 to 5. }
    \label{Fig:chatgpt_sentiment}
    \end{figure}

    \item \label{point:model_fine_tuning} Alternative models and model fine-tuning: The majority of available pre-trained language models that have the ability to classify text by sentiment, have not been developed for the medical or health domains as their targets. This likely to change in the near future. Note \ref{point:alternative_models} directs the researchers to a main repository for the transformer-based models where alternatives and new models can be located. 

    The possibility also remains for researchers to customise underlying transformer-based models such as BERT for the medical and health domains in order to enhance the performance of sentiment models. This is an optional step and is not covered in this chapter. This process would require a certain level of technical skill as well as a commitment of time to create datasets that can fine-tune the underlying BERT model, but would however be a valuable contribution.

\end{enumerate}

\section{Notes}\label{notes}

\begin{enumerate}
\item \label{point:installation} Software installation: Anaconda can be freely downloaded from \url{https://www.anaconda.com/products/distribution}. Deepnote  (\url{https://deepnote.com}) is an alternative cloud-based provider of scientific Python libraries and has a user-friendly programming interface. The study was conducted with the installation of two additional packages shown in Fig. \ref{Fig:install}.

    \begin{figure}[hbt]
    \centering
    \includegraphics[scale=0.4]{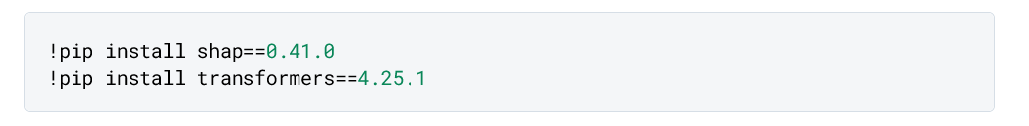}
    \caption{Required Python packages to be installed: Transformers \cite{wolf-etal-2020-transformers} and SHAP. }
    \label{Fig:install}
    \end{figure}

\item \label{point:model_details} Information on the BERT model used in this demonstration can be found here: \url{https://huggingface.co/nlptown/bert-base-multilingual-uncased-sentiment}. A current limitation of the specific BERT models used in this demonstration is that it accepts a maximum of 512 tokens (words) at a time. Since some abstracts are longer than this, and to get around this limitation, abstracts can be divided into smaller chunks and a weighted average across all chunks can then be calculated in order to assign the sentiment value to the whole abstract.

\item \label{point:tokenisation} Tokenisation is one additional step that is usually performed before most sentiment analysis models can be used that rely on standard machine learning models or lexicon-based approaches.

\item \label{point:domain_mismatch} Conducting sentiment analysis on the scientific literature on the topic of Lyme disease poses two significant difficulties. The first is that scientific literature strives to present results in a factual and objective manner, without expressing any positive or negative sentiment, making it difficult for models to detect subtleties in sentiment. Secondly, the difficulty is compounded by the fact that most pre-trained models have been trained on text dataset from the business context comprising customer reviews. The challenge here is that certain domain-specific terms like 'pathology' in a medical context may have a neutral sentiment, while in a business context, this word may have a negative connotation. A mismatch between the source domain of a pre-trained model and the target domain is a considerable challenge and is an area of ongoing research \cite{deepa2021bidirectional}. Research in cross-domain sentiment classification and transfer learning are areas of focus which aim to address this challenge.

\item \label{point:alternative_models} A list of alternative and fine-tuned sentiment analysis transformer-based models can be found here \url{https://huggingface.co/models?sort=downloads&search=sentiment}. In time, domain-specific models tailored for the medical and health research domain will become available. Some models for the clinical domain already exist but they do not have the capability to perform sentiment analysis.

\end{enumerate}

\section{Acknowledgement}\label{sec1}

The author would like to extend his sincere gratitude to Jamie Scott for his invaluable contributions in collecting the dataset used in this work as well as for the data cleaning and preliminary analyses. His dedication and hard work have been instrumental in making this project possible. Additionally, the author thanks Professor Stephen Croucher for initially proposing the idea of examining sentiment in chronic Lyme disease literature. 

\begin{multicols}{2}


\begin{thebibliography}{8}
\providecommand{\natexlab}[1]{#1}
\providecommand{\url}[1]{\texttt{#1}}
\expandafter\ifx\csname urlstyle\endcsname\relax
  \providecommand{\doi}[1]{doi: #1}\else
  \providecommand{\doi}{doi: \begingroup \urlstyle{rm}\Url}\fi

\bibitem[Deepa et~al.(2021)]{deepa2021bidirectional}
M.~D. Deepa et~al.
\newblock Bidirectional encoder representations from transformers (bert)
  language model for sentiment analysis task.
\newblock \emph{Turkish Journal of Computer and Mathematics Education
  (TURCOMAT)}, 12\penalty0 (7):\penalty0 1708--1721, 2021.

\bibitem[Devlin et~al.(2018)Devlin, Chang, Lee, and Toutanova]{devlin2018bert}
J.~Devlin, M.-W. Chang, K.~Lee, and K.~Toutanova.
\newblock Bert: Pre-training of deep bidirectional transformers for language
  understanding.
\newblock In \emph{Proceedings of the 2018 Conference of the North American
  Chapter of the Association for Computational Linguistics: Human Language
  Technologies}, volume~1, pages 4171--4186, 2018.

\bibitem[Halperin(2015)]{halperin2015chronic}
J.~J. Halperin.
\newblock Chronic lyme disease: misconceptions and challenges for patient
  management.
\newblock \emph{Infection and Drug Resistance}, pages 119--128, 2015.

\bibitem[Levesque and Klohn(2019)]{levesque2019multiple}
M.~Levesque and M.~Klohn.
\newblock A multiple streams approach to understanding the issues and
  challenges of lyme disease management in canada’s maritime provinces.
\newblock \emph{International Journal of Environmental Research and Public
  Health}, 16\penalty0 (9):\penalty0 1531, 2019.

\bibitem[Lundberg and Lee(2017)]{NIPS2017_7062}
S.~M. Lundberg and S.-I. Lee.
\newblock A unified approach to interpreting model predictions.
\newblock In I.~Guyon, U.~V. Luxburg, S.~Bengio, H.~Wallach, R.~Fergus,
  S.~Vishwanathan, and R.~Garnett, editors, \emph{Advances in Neural
  Information Processing Systems 30}, pages 4765--4774. Curran Associates,
  Inc., 2017.

\bibitem[Rebman and Aucott(2020)]{rebman2020post}
A.~W. Rebman and J.~N. Aucott.
\newblock Post-treatment lyme disease as a model for persistent symptoms in
  lyme disease.
\newblock \emph{Frontiers in medicine}, page~57, 2020.

\bibitem[Wolf et~al.(2020)Wolf, Debut, Sanh, Chaumond, Delangue, Moi, Cistac,
  Rault, Louf, Funtowicz, Davison, Shleifer, von Platen, Ma, Jernite, Plu, Xu,
  Scao, Gugger, Drame, Lhoest, and Rush]{wolf-etal-2020-transformers}
T.~Wolf, L.~Debut, V.~Sanh, J.~Chaumond, C.~Delangue, A.~Moi, P.~Cistac,
  T.~Rault, R.~Louf, M.~Funtowicz, J.~Davison, S.~Shleifer, P.~von Platen,
  C.~Ma, Y.~Jernite, J.~Plu, C.~Xu, T.~L. Scao, S.~Gugger, M.~Drame, Q.~Lhoest,
  and A.~M. Rush.
\newblock Transformers: State-of-the-art natural language processing.
\newblock In \emph{Proceedings of the 2020 Conference on Empirical Methods in
  Natural Language Processing: System Demonstrations}, pages 38--45, Online,
  Oct. 2020. Association for Computational Linguistics.
\newblock URL \url{https://www.aclweb.org/anthology/2020.emnlp-demos.6}.

\bibitem[Wong et~al.(2022)Wong, Shapiro, and Soffer]{wong2022review}
K.~H. Wong, E.~D. Shapiro, and G.~K. Soffer.
\newblock A review of post-treatment lyme disease syndrome and chronic lyme
  disease for the practicing immunologist.
\newblock \emph{Clin Rev Allergy Immunol}, 62\penalty0 (1):\penalty0 264--271,
  2022.

\end{thebibliography}
\end{multicols}

\end{document}